\documentclass[conference]{IEEEtran}
\IEEEoverridecommandlockouts
\usepackage{cite}
\usepackage{amsmath,amssymb,amsfonts}
\usepackage{algorithmic}
\usepackage{graphicx}
\usepackage{textcomp}
\usepackage{xcolor}
\usepackage{url}
\usepackage{natbib}
\usepackage{graphicx}
\usepackage{booktabs} 
\usepackage{subcaption}
\usepackage{hyperref}
\def\BibTeX{{\rm B\kern-.05em{\sc i\kern-.025em b}\kern-.08em
    T\kern-.1667em\lower.7ex\hbox{E}\kern-.125emX}}
\begin{document}

\title{A Comparative Study of NAFNet Baselines for Image Restoration\\
}

\author{
\IEEEauthorblockN{Vladislav Esaulov\IEEEauthorrefmark{1} \hspace{2cm} M. Moein Esfahani\IEEEauthorrefmark{2}} 
\IEEEauthorblockA{\IEEEauthorrefmark{1}Department of Computer Science, Georgia State University, Atlanta, USA\\
Email: vesaulov1@gsu.edu}
\IEEEauthorblockA{\IEEEauthorrefmark{2}Center for Translational Research in Neuroimaging and Data Science (TReNDS), Georgia State University, Atlanta, USA\\
Email: mesfahani1@gsu.edu}
}

\maketitle

\begin{abstract}
We study NAFNet (Nonlinear Activation Free Network), a simple and efficient deep learning baseline for image restoration. By using CIFAR10 images corrupted with noise and blur, we conduct an ablation study of NAFNet's core components. Our baseline model implements SimpleGate activation, Simplified Channel Activation (SCA), and LayerNormalization. We compare this baseline to different variants that replace or remove components. Quantitative results (PSNR, SSIM) and examples illustrate how each modification affects restoration performance. Our findings support the NAFNet design: the SimpleGate and simplified attention mechanisms yield better results than conventional activations and attention, while LayerNorm proves to be important for stable training. We conclude with recommendations for model design, discuss potential improvements, and future work.
\end{abstract}


The source code is available on GitHub at \href{https://github.com/vlad777442/exploring-nafnet-baselines}{this repository}.

\section{Introduction}

\subsection{Image Restoration}

Image restoration aims to effectively recover high-quality images from degraded inputs (e.g., noisy or blurry images). It's important for low-level vision tasks with applications in photography, medical images, and surveillance. Recent deep-learning models have achieved state-of-the-art quality, but often they consist of complex architectures and heavy attention mechanisms. For example, multi-stage networks and transformer-based networks can achieve high PSNR, but they are computationally expensive. This growing complexity can hinder insight on which architectural components are really important. 

\subsection{NAFNet}

Chen et al. (2022) \cite{chen2022simple} addresses this by proposing a "simple baseline" architecture for image denoising that reduces complexity while retaining high performance. This baseline is built on a UNet backbone (single-stage, encoder-decoder) to ensure low inter-block complexity. For intra-block design, they incrementally added common enhancements (normalization, advanced activation, attention) to a plain CNN block and measured the improvement. Through extensive ablations, they converged on a minimalist yet effective design called NAFNet (Nonlinear Activation Free Network). NAFNet's key insight is that standard nonlinearities (ReLU, GELU, Sigmoid, etc.) can be replaced by simple linear gating and attention modules. Their model achieves competitive PSNR on deblurring (GoPro) and denoising (SIDD) benchmarks while using a fraction of the computational resources of prior models.

\subsection{Motivation}

We aim to reproduce and analyze NAFNet's baseline design on a smaller-scale task. We focus on CIFAR10 images (32x32) with synthetic blur and noise. This allows us to validate its design and principles and understand the role of each component. In particular, we explore replacing NAFNET's custom modules with more conventional choices and see how this affects restoration quality. How important are NAFNet's components, and can further simplifications maintain or improve results?

\subsection{Contributions}

We implement a CIFAR10 image restoration pipeline based on NAFNet's architecture and conduct several ablation experiments:

\begin{enumerate}
    \item \textbf{Baseline.} A NAFNet block with SimpleGate + SCA + LayerNorm.
    \item \textbf{A1 (Activation).} Replace the SimpleGate with a standard GELU activation.
    \item \textbf{A2 (Attention).} Replace the Simplified Channel Attention with an Efficient Channel Attention (ECA) module.
    \item \textbf{A3 (Normalization).} Replace Layer Normalization with Group Normalization.
    \item \textbf{A4.} Remove the attention module entirely (no SCA/ECA).
\end{enumerate}

We report comparative performance of these variants using PSNR and SSIM \cite{wang2004ssim}. We also provide visual examples to highlight qualitative differences. Our findings largely align with NAFNet paper - the simplified gating and attention are surprisingly effective - while offering lightweight image restoration models. Finally, we suggest improvements and future work for extending this study. 

\section{Related Work}

\subsection{Image Restoration}

Traditional methods for deblurring and denoising images often rely on optimization (e.g. deconvolution, BM3D \cite{dabov2007bm3d}), while modern solutions leverage deep learning for mappings from degraded to clean images. High-performing models include multi-stage networks (e.g. MPRNet \cite{zamir2021mprnet}) and transformer-based architectures (e.g. SwinIR \cite{liang2021swinir}, Restormer  \cite{zamir2022restormer}) which introduce complex self-attention blocks. While effective, these methods have high system complexity - many layers, branches, or heavy per-block computations. This makes them computationally expensive and hard to interpret.

\subsection{NAFNet Baseline}

The work "Simple Baselines for Image Restoration" challenges the need of such complexity. The authors first formulate a PlainNet consisting of just convolutions, ReLU, and skip connections - a simple residual CNN block. Then, they incrementally add modules:

\begin{itemize}
    \item \textbf{Layer Normalization (LN):} to improve training stability and performance. It normalizes feature across channels/spatial for each sample. Chen et al. (2022) \cite{chen2022simple} found adding LN to plain block gave significant gains (+0.44 dB on SIDD denoising, +3.39 dB on GoPro deblurring), stabilizing training where batch norm would fail for small batches.
    \item \textbf{GELU Activation:} replacing ReLU with GELU yielded a minor improvement on deblurring (+0.21 dB) while maintaining performance on denoising.
    \item \textbf{Channel Attention (CA):} a SENet-style \cite{hu2018squeeze} channel attention module that weighs feature channels by their global importance. This added a 0.1-0.2 dB boost in tests.
\end{itemize}

Combining these, their baseline block has LN, conv layers, GELU, and CA. Moreover, each component individually is "trivial," yet together they surpassed prior SOTA models in quality.

The authors further simplified the baseline by observing that GELU is essentially a gated activation - it can be viewed as $x \times \Phi(x)$, where $\Phi$ is a non-linear function. They propose to replace GELU with a SimpleGate (SG): split the feature channels in two and multiply them element-wise. This gating has no explicit non-linearity. Replacing GELU with SimpleGate actually improved PSNR by a small margin in their experiments, showing that learnable gating can fully substitute for GELU's effect. Similarly, they simplify the Channel Attention: the standard CA uses a sigmoid activation and an MLP to generate channel weights; NAFNet's Simplified Channel Attention removes all non-linearities from CA. In practice, SCA computes channels via a single linear transformation of global pooled features and uses no sigmoid - effectively allowing weights $>1$ or negative, which network can calibrate. Impressively, NAFNet has no ReLU/GELU/Sigmoid at all. It also matched the baseline performance (e.g. 39.96 vs 39.85 dB on SIDD) while boosting inference by $9\%$.

\subsection{Efficient Channel Attention (ECA):} 
In one of our experiments, we consider ECA as an alternative attention mechanism. ECA is a lightweight channel attention that avoids dimensionality reduction of CA by using a 1D convolution on the channel-wise global pool vector, followed by a sigmoid. It intoduces minimal overhead and has been shown to be effective in classification tasks with less parameters than SE/CA. We use ECA to re-introduce a non-linear activation, but it's useful to see if added non-linearity and a different weighting strategy affect performance.

\subsection{LayerNorm vs GroupNorm} 
NAFNet uses LayerNorm in each block, due to its widespread use in Transformers and stable behavior even with small batches. Group Normalization \cite{wu2018groupnorm} is another mechanism that normalizes channels in groups and does not depend on batch statistics. GN is successful in vision tasks when batch norm fails. Our experiment replaces LN with GN to test if simpler normalization can perform similarly, Prior work suggests LN helps to preserve feature scale in restoration transformers.

\section{Materials and Methods}

\begin{figure}[!t]
    \centering
    \includegraphics[width=\columnwidth]{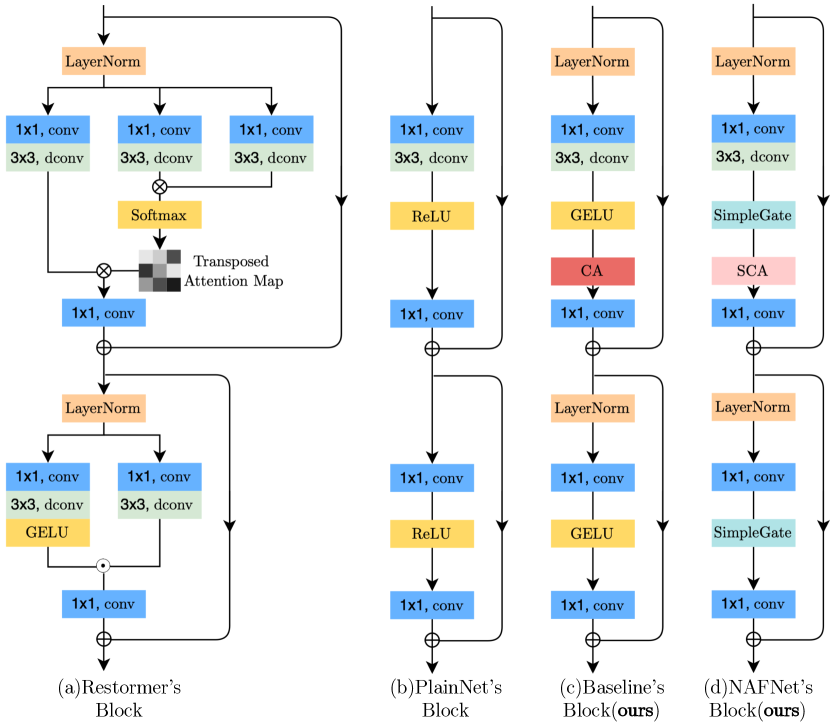}
    \caption{Illustration of different residual block designs (adapted from \cite{chen2022simple}). (a) Restormer’s transformer-based block (for reference, very complex). (b) Plain CNN block with conv + ReLU (no normalization or attention). (c) Improved baseline block with LayerNorm, GELU activation, and Channel Attention (CA). (d) NAFNet block with LayerNorm, SimpleGate (SG) instead of GELU, and Simplified Channel Attention (SCA) instead of CA.}
    \label{fig:figure1}
\end{figure}

\subsection{Network Architecture}

The baseline follows NAFNet's simple single-stage UNet design. The network has an encoder-decoder structure with skip connections, using NAF blocks in each stage instead of more complex transformers or dense blocks. UNet is pretty efficient in image-to-image tasks and aligns with NAFNet baseline assumptions. Each NAF block in the network is a residual unit:

\begin{enumerate}
    \item \textbf{LayerNorm:} normalize the input feature map. 
    \item \textbf{1x1 convolution:} a pointwise convolution to mix channel information and change dimensions as needed. 
    \item \textbf{3x3 depthwise convolution:} a spatial convolution to process local neighborhood information.
    \item \textbf{Activation:} GELU or SimpleGate to introduce nonlinearity 
    \item \textbf{1x1 convolution:} pointwise conv to produce the block's output features. 
    \item \textbf{Channel Attention:} SCA or ECA, this module takes the output and recalibrates channel magnitudes.
    \item \textbf{Residual Attention:} the block's input is added back to the output to form the final block.
\end{enumerate}

We use PyTorch in order to implement the baseline, matching the open-source NAFNET where possible. The baseline uses SimpleGate for the activation and SCA for attention, with LayerNorm. The network depth and feature channels were adjusted for CIFAR10 - we use a smaller model. Figure \ref{fig:figure1} shows four models, yellow boxes denote nonlinear activations – notably, block (d) has none, and it uses SG and SCA to achieve a nonlinear effect without explicit activation functions. We use (d) as our baseline block design.

\subsection{Experimental Plan}

To understand the contribution of each component, we decided to test four models, each differing from the baseline in some aspects.

\begin{itemize}
    \item \textbf{A1: Replace SimpleGate with GELU.} In this experiment we test the importance of NAFNet's gating mechanism. We hypothesize that if SimpleGate is crucial, using GELU will decrease performance. The authors reported that switching from GELU to SimpleGate boosted PSNR slightly, so we expect the baseline to have a small edge over A1. 
    \item \textbf{A2: Replace SCA with ECA.} This experiment reintroduces a conventional attention in place of the simplified linear attention. ECA is computationally lightweight, so this swap keeps the model efficient. We want to see if the nonlinearity in channel attention yields any gain. If performance increases it may indicate that some nonlinear modulation of channels is beneficial.
    \item \textbf{A3: Replace LayerNorm with GroupNorm.} Here we examine normalization. LayerNorm operates across all channels for each pixel and is being widely used in transformer-based networks. GroupNorm normalizes within a smaller group of channels. We expect minor differences: LN may yield slightly higher SSIM/PSNR due to better preservation of features, while GN could perform similarly if configured well.
    \item \textbf{A4: Remove Attention.} We disable the attention module completely, the block ends with the last 1x1 conv and adds the residual skip without channel re-weighting. This shows how much any attention contributes. Prior studies showed that removing attention led to a noticeable drop. 
\end{itemize}

All other aspects (architecture, parameters) are kept constant across experiments so that comparisons isolate the effect of these changes. We use the same number of NAF blocks and channel dimensions as in the baseline. This approach follows the methodology of the NAFNet paper.

\subsection{Dataset}

We use CIFAR10 dataset \cite{krizhevsky2009learning} as the source of images. This dataset consists of 60000 images (32x32 pixels) across 10 classes. We chose this dataset because its low resolution makes the restoration task non-trivial. We set 50000 images for training and 10000 for testing, using the standard split. 

Each image is degraded with a combination of blur and noise to create input-output pairs for training. 

\begin{figure}[!t]
    \centering
    \includegraphics[width=0.45\columnwidth]{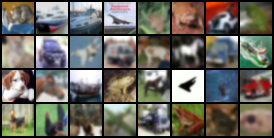}
    \hfill
    \includegraphics[width=0.45\columnwidth]{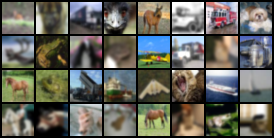}
    \hfill
    \includegraphics[width=0.45\columnwidth]{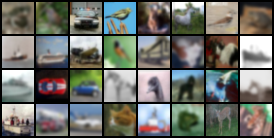}
    \hfill
    \includegraphics[width=0.45\columnwidth]{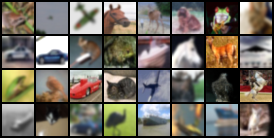}
    % \hfill
    % \includegraphics[width=0.3\columnwidth]{a3_noisy_4.png}
    \caption{Degraded images}
    \label{fig:figure2}
\end{figure}

\begin{itemize}
    \item \textbf{Blur.} We apply a Gaussian blur with a random sigma in the range $[0, 3]$. This simulates out-of-focus or motion blur.
    \item \textbf{Noise.} Gaussian noise with a random sigma in $[0, 30]$. This produces a noisy effect like low-light conditions. The degradations are applied uniformly across the dataset to make sure there are no dataset leak. An example of images is shown in Figure \ref{fig:figure2}.
\end{itemize}

\subsection{Training details}

We train each model from scratch on the training dataset. We use Mean Squared Error (MSE) as the loss, and the Adam optimizer. We use a batch size of 16 and train for 50 epochs. The training also employs learning rate scheduling and early stopping if validation stops improving. 

\subsection{Evaluation Metrics}

We evaluate results with:

\begin{itemize}
    \item \textbf{PSNR (Peak Signal-to-Noise Ratio).} Measures the reconstruction fidelity in dB. Higher PSNR indicates the restored image is closer to ground truth.
    \item \textbf{SSIM (Structural Similarity Index).} Measures perceptual image quality (0 to 1). We compute the mean SSIM over the test set for each model, as a complement to PSNR, since SSIM better captures structural distortion and human perceptual quality.
    \item \textbf{LPIPS (Learned Perceptual Image Patch Similarity).} It measures perceptual similarity (closer to human judgment) \cite{zhang2018lpips}. Shows how visually realistic outputs look, even if PSNR/SSIM are close.
\end{itemize}

\section{Results}

Table~\ref{tab:ablation} summarizes the performance of the baseline and each variant on the CIFAR10 dataset. We include PSNR (in dB, higher is better), SSIM (higher is better), and LPIPS (a lower score indicates higher perceptual similarity). Bold indicates the best value in each column.

\begin{table}[h]
\centering
\caption{Results of NAFNet modifications.}
\label{tab:ablation}
\begin{tabular}{lccc}
\toprule
\textbf{Variant} & \textbf{PSNR (dB)} & \textbf{SSIM} & \textbf{LPIPS} $\downarrow$ \\
\midrule
Baseline & 29.37 & \textbf{0.9565} & 0.0073 \\
A1: Replace SimpleGate $\rightarrow$ GELU & 29.14 & 0.9550 & 0.0082 \\
A2: Replace SCA $\rightarrow$ ECA & 28.86 & 0.9556 & 0.0093 \\
A3: Replace LayerNorm $\rightarrow$ GroupNorm & \textbf{29.38} & 0.9562 & 0.0079 \\
A4: Remove Attention & 28.80 & 0.9535 & 0.0094 \\
\bottomrule
\end{tabular}
\end{table}

\begin{figure}[!t]
    \centering
    \begin{subfigure}{0.45\columnwidth}
        \centering
        \includegraphics[width=\linewidth]{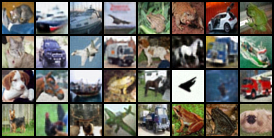}
        \caption{Baseline model output}
        \label{fig:baseline1}
    \end{subfigure}
    \hfill
    \begin{subfigure}{0.45\columnwidth}
        \centering
        \includegraphics[width=\linewidth]{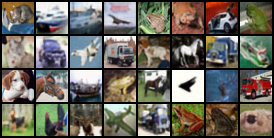}
        \caption{A1 model output}
        \label{fig:baseline2}
    \end{subfigure}
    \hfill
    \begin{subfigure}{0.45\columnwidth}
        \centering
        \includegraphics[width=\linewidth]{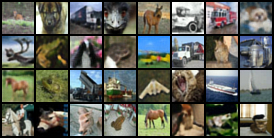}
        \caption{A2 model output}
        \label{fig:baseline3}
    \end{subfigure}
    \hfill
    \begin{subfigure}{0.45\columnwidth}
        \centering
        \includegraphics[width=\linewidth]{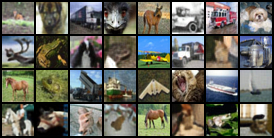}
        \caption{A3 model output}
        \label{fig:a3}
    \end{subfigure}
    \caption{Comparison of denoised outputs: (a) Baseline model, (b) A1 model, (c) A2 model, (d) A3 model.}
    \label{fig:comparison}
\end{figure}

One of the NAFNet's claims is achieving strong performance with lower complexity. While our experiments were conducted on a smaller scale, their claim holds true. To support our quantitative comparisons, we provide models output in Figure~\ref{fig:comparison}. In this figure the baseline model (a) delivers a sharp and clean restoration. Noise is almost completely removed, edges are well-reconstructed, and colors closely match the ground truth. The A1 model (b), which replaces SimpleGate with GELU, produces an image very similar to the baseline, but shows slightly more residual noise and minor edge softness. The A3 model showed pretty much similar results to the baseline.

Overall, qualitative results align with our numerical findings:

\begin{itemize}
    \item Removing attention notably hurts texture recovery and fine detail.
    \item Replacing SimpleGate with GELU has subtle but measurable effects.
    \item And the baseline design delivers the best balance between denoising, deblurring, and preserving structural information.
\end{itemize}

\section{Discussion}

One of the NAFNet's core goals was to achieve high restoration quality, while maintaining low complexity in comparison to other deep learning models. Our results reaffirm these findings. By systematically modifying the activation function, attention mechanism, and normalization layer, we observed how each design choice impacted image restoration performance. Below, we analyze the outcomes and offer possible explanations for the observed trends.

\begin{itemize}
    \item A3 (GroupNorm) slightly outperformed the baseline in terms of PSNR and had comparable SSIM and LPIPS. GroupNorm may offer better localized normalization for small spatial dimensions (32×32), as it operates across grouped channels instead of normalizing across the entire layer like LayerNorm. This can preserve spatial information more effectively in low-resolution images. However, the margin was small, indicating both normalization schemes are suitable for this scale.
    \item The baseline model achieved a strong balance across all three metrics. It retained the original NAFNet structure: SimpleGate activation, Simplified Channel Attention, and LayerNorm. This confirms that NAFNet’s simplified design generalizes well to different datasets—even small, without the need for complex modules.
    \item A1 (GELU instead of SimpleGate) produced results close to the baseline, with slightly lower PSNR and higher LPIPS. While GELU is a standard, well-behaved activation, it lacks the input-adaptive modulation that SimpleGate introduces. SimpleGate multiplies two feature maps, offering dynamic gating without adding heavy computation. This difference in flexibility may explain why GELU underperforms slightly.
    \item A2 (ECA instead of SCA) degraded performance more than expected. Although ECA is widely used for channel attention and introduces only a lightweight 1D convolution followed by a sigmoid, its nonlinearity might interfere with the model’s ability to balance intensity. In contrast, SCA uses a purely linear weighting, which may retain more precise feature scaling needed for pixel-wise regression. 
\end{itemize}

\section{Conclusion and Future Work}

In this project, we conducted a thorough architecture ablation study on a simple NAFNet-based network for image restoration on CIFAR10 with synthetic blur and noise.
Through systematic experiments, we validated that NAFNet’s design—favoring SimpleGate activations, Simplified Channel Attention, and LayerNorm—indeed strikes an excellent balance between performance and simplicity.

\begin{itemize}
    \item SimpleGate slightly outperformed GELU for restoration, supporting the use of learnable gating over traditional nonlinearities.
    \item SCA proved more effective than ECA, showing that nonlinearity in attention is not strictly necessary for strong performance on small images.
    \item GroupNorm surprisingly achieved the highest PSNR, suggesting that for small-scale image restoration tasks, GroupNorm can be a viable and even superior to LayerNorm.
    \item Attention modules are critical — removing them caused clear degradation, emphasizing the importance of global feature modulation.
\end{itemize}

Overall, these results reinforce the idea that carefully streamlined architectures can deliver competitive results with fewer parameters and reduced complexity, which is highly desirable for real-time or resource-constrained applications.

Future work includes:

\begin{itemize}
    \item Benchmark on Higher-Resolution Data: Test whether these conclusions hold on larger images (e.g., BSD500, GoPro, or SIDD datasets).
    \item Alternative Normalizations: Evaluate InstanceNorm or spatially-adaptive normalizations to potentially improve restoration consistency.
    \item Enhanced Training Strategies: Introduce perceptual or adversarial losses to further improve visual realism to target lower LPIPS scores.
\end{itemize}

\bibliographystyle{plain} % or another valid style
\bibliography{references} % matches the filename of your .bib file (no .bib extension)

\vspace{12pt}

\end{document}